\documentclass[letterpaper, 10 pt, conference]{ieeeconf}  

\IEEEoverridecommandlockouts                              
\usepackage[pdftex]{graphicx}
\usepackage{amssymb}
\usepackage{array}
\usepackage{bm}
\usepackage[]{units}
\usepackage{siunitx}
\usepackage{url}
\usepackage{color}
\usepackage{epstopdf}
\usepackage{mathtools}
\usepackage{amsmath}

\usepackage{amsthm}
\usepackage[export]{adjustbox}
\usepackage[ruled]{algorithm2e} 
\usepackage{cite}

\newtheorem{definition}{Definition}

\makeatletter
\let\NAT@parse\undefined
\makeatother
\usepackage{hyperref}  

\title{\LARGE \bf
MPC-Based Hierarchical Task Space Control of Underactuated and Constrained Robots for Execution of Multiple Tasks}

\author{Jaemin Lee$^{1}$, Seung Hyeon Bang$^{2}$, Efstathios Bakolas$^{2}$, and Luis Sentis$^{2}$
\thanks{This work was supported by NSF Grant\# 1724360 and ONR Grant\# N000141512507. The third author acknowledges partial support by NSF award ECCS-1924790.}
\thanks{$^{1}$J. Lee, is with the Department of Mechanical Engineering, The University of Texas at Austin, TX 78712, USA
                {\tt\small jmlee87@utexas.edu}}%
\thanks{$^{2}$S.H. Bang, E. Bakolas, and L. Sentis are with the Department of Aerospace Engineering and Engineering Mechanics, The University of Texas at Austin, TX 78712, USA
        {\tt\small bangsh0718@utexas.edu, \{bakolas, lsentis\}@austin.utexas.edu}}%
}

\begin{document}

\maketitle
\thispagestyle{empty}
\pagestyle{empty}

\begin{abstract}
This paper proposes an MPC-based controller to efficiently execute multiple hierarchical tasks for underactuated and constrained robotic systems. Existing task-space controllers or whole-body controllers solve instantaneous optimization problems given task trajectories and the robot plant dynamics. However, the task-space control method we propose here relies on the prediction of future state trajectories and the corresponding costs-to-go terms over a finite time-horizon for computing control commands. We employ acceleration energy error as the performance index for the optimization problem and extend it over the finite-time horizon of our MPC. Our approach employs quadratically constrained quadratic programming, which includes quadratic constraints to handle multiple hierarchical tasks, and is computationally more efficient than nonlinear MPC-based approaches that rely on nonlinear programming. We validate our approach using numerical simulations of a new type of robot manipulator system, which contains underactuated and constrained mechanical structures.
\end{abstract}

\section{Introduction}
Highly articulated robots are increasingly employed and rely on hierarchical task execution to operate in dynamic environments. The Operational Space Control (OSC) method considers manipulators' end-effector dynamics, virtually decoupling closed-loop task-dynamics in the end-effector's operational space \cite{khatib1987unified}. In particular, OSC computes dynamically consistent torque commands needed for robots to effectively track motion and force trajectories. A detailed analysis of OSC for constrained and underactuated robotic systems is presented in \cite{mistry2012operational}. In recent times, the OSC method has been extended to the Whole Body Control (WBC) method which incorporates floating base robots \cite{sentis2005control}, multi-contact, and dynamically consistent constraints, task and posture primitives \cite{sentis2006whole}. WBC has been broadly applied to bipedal humanoid robots \cite{sentis2010compliant,gienger2005task,kim2016stabilizing,lee2016balancing} and mobile humanoid robots \cite{dietrich2012reactive}. In the case of bipedal humanoids, the WBC method has been employed to achieve multi-contact balance behaviors \cite{sentis2007synthesis,sentis2010compliant} and various complex motions such as dynamically walking, jumping, climbing up a ladder \cite{park2006contact}, and hand manipulation \cite{mansard2009unified}. Several other relevant whole-body control techniques have been proposed such as \cite{ott2011posture}, which relies on the manipulation of contact forces on the ground and multi-contact points.

Although OSC and WBC are capable of computing control commands fast and establishing a real-time feedback loop, they are both based on instantaneous optimization, e.g., least-square error minimization. For this reason, the control command is only optimal locally at each time instance. Model Predictive Control (MPC) offers an alternative in which an optimal control problem can be recursively solved over a finite-time horizon \cite{camacho2013model, grune2017nonlinear}. MPC has been employed successfully for the control of robotic manipulators \cite{wahrburg2016mpc,ferrara2013robust}. Nonlinear MPC (NMPC) is employed for the control of manipulators in the presence of external disturbances \cite{poignet2000nonlinear}, for image-based visual servoing with visibility constraints \cite{sauvee2006image}, and as a robust control strategy \cite{hajiloo2015robust}. Also, MPC solves practical problems in robotics such as collision avoidance \cite{eskandarpour2014cooperative} or singularity avoidance \cite{nikou2017nonlinear} over a finite time horizon. In addition, whole body MPC is achieved using Sequential Linear Quadratic (SLQ) programming for mobile manipulators \cite{minniti2019whole}. In the case of locomotion for quadruped robots, MPC has been employed to find contact forces that allow a lumped mass model of the robot to track desired trajectories obtained via WBC \cite{kim2019highly}.

Two significant issues arise when trying to replace the conventional OSC or WBC with MPC: 1) Dealing with the whole-body nonlinear dynamics, and 2) Dealing with the task hierarchy commonly imposed in highly articulated robots performing multiple tasks. Most robotic systems are nonlinear, constrained, and sometimes underactuated. Often, simplified models such as a linearized CoM model of a walking robot are employed to approximate contact forces during locomotion and used in MPC structures \cite{kim2019highly}. However, such methods employ an additional WBC step to generate actuator commands. In our approach, we remove the need to rely on simplified models for control and directly linearize complex multi-body models of robotic systems to reduce the computation time of MPC \cite{gros2020linear}. The main reason why robots have been employing OSC and WBC methods is that they can be simply executed using a single Quadratic Programming (QP) optimization step, which is substantially faster than employing MPC or NMPC. Another reason why OSC and WBC have become popular is because they can compute control commands achieving multiple task goals and organized as a hierarchy, for instance using projection-based methods \cite{lee2012intermediate} or Hierarchical QP (HQP) \cite{escande2014hierarchical} at each time instance. To achieve better control performance over a finite-time horizon, we propose to transform OSC and WBC into a convex MPC while fulfilling multiple task goals and constraints as required for control of complex robotic systems. 

Next, we summarize the main contributions of our work. We formulate a new MPC as a transformation of WBC and OSC for effective control of underactuated and constrained robots. To the best of our knowledge, this is the first study to propose an MPC-based extension of WBC for the execution of hierarchical tasks. We linearize the nonlinear robot dynamics with respect to nominal joint space trajectories obtained via Inverse Kinematics (IK) or Inverse Dynamics (ID) operations applied to previously defined task trajectories. The running cost for the proposed MPC is constructed to mimic the optimization cost associated with WBC and OSC. Furthermore, we classify the task hierarchy as being either a \textit{weak hierarchy} or a \textit{strong hierarchy} each associated with a corresponding quadratic inequality constraint. \textit{Weak hierarchy} implies that the tracking error of higher prioritized task must be equal or smaller than that of lower prioritized task. \textit{Strong hierarchy} of multiple tasks impose the constraints that the higher prioritized task error has to be strictly smaller than that of lower prioritized task. In our MPC-based approach, the cost and constraint functions associated with the execution of the hierarchical tasks are convex quadratic functions and the system dynamics are linearized as previously mentioned; thus, each finite-horizon optimal control problem can be associated with a Quadratically Constrained Quadratic Program (QCQP), which can be solved using convex optimization tools. 

For validation, we apply the proposed convex MPC-based approach to \textit{Scorpio}, a unique robotic manipulator equipped with $7$ moving DOF, where two of them are implemented using mechanical parallelograms corresponding to movement elevations. Although the robotic manipulator is able to handle high-payload objects effectively due to the distinctive mechanisms, the control problem of the robotic system becomes more complicated. More specifically, in each parallelogram, there exist one driving joint, two passive joints, and one kinematic constraint. We validate the proposed QCQP-based MPC by demonstrating numerical simulations of this robot \textit{Scorpio}, while the results are compared with the behavior resulting from using a simpler WBC controller.

The remainder of this paper is organized as follows. We briefly review WBC for underactuated and constrained robots in Section \ref{sec:2}. In Section \ref{sec3}, we present the proposed MPC-based approach and explain its implementation via convex optimization tools. In Section \ref{sec4}, we apply the proposed methodology to \textit{Scorpio}, which is a unique underactuated and constrained manipulator. Numerical simulations are also provided to show the effectiveness of the proposed control method. 

\section{Preliminaries}
\label{sec:2}

\subsection{Notation}
We represent the sets of $n$ dimensional real vectors and $m \times n$ matrices by $\mathbb{R}^{n}$ and $\mathbb{R}^{m\times n}$, respectively. $\mathbb{S}_{+}^{n}$ and $\mathbb{S}_{++}^{n}$ denote the sets of $n \times n$ positive semi-definite and positive definite matrices, respectively. Given $n$ real numbers $a_1, \cdots, a_n$, $\textrm{diag}(a_1, \cdots, a_n)$ represents the $n\times n$ matrix whose diagonal terms are $a_1, \cdots, a_n$. $\textrm{bdiag}(\mathbf{A}_1, \cdots, \mathbf{A}_n)$ denotes the block diagonal matrix constructed by matrices $\mathbf{A}_1, \cdots, \mathbf{A}_n$ of compatible dimensions. $\mathbf{A}^{\dag}$ denotes the Moore-Penrose pseudo inverse of $\mathbf{A}$, which is a real matrix. In addition, we express a discretized interval of $\left[ a, \: b\right]$ as $\left[a, \: b\right]_{d}$ where $a$ and $b$ are integers with $a\leq b$. Finally, $\mathbf{1}_{n} \in \mathbb{R}^{n}$ denotes the $n$-dimensional vector whose components are all equal to $1$.

\subsection{Whole Body Controller}
The rigid body dynamics equation for $n$ DOF robots actuated by $m$ joints ($m\leq n$) is expressed as follows:
\begin{equation}
    \mathbf{M}(q) \ddot{q}  + b\left(\dot{q}, q\right) +  \mathbf{J}_{c}^{\top}(q) F_{c} = \mathbf{U}^{\top} \Gamma
\label{eq:dynamics}
\end{equation}
where $q\in \mathbb{R}^{n}$, $\mathbf{M}(q)\in \mathbb{R}^{n \times n}$, $b(q,\dot{q}) \in \mathbb{R}^{n}$, $\mathbf{J}_{c}(q) \in \mathbb{R}^{n_c \times n}$, $F_{c} \in \mathbb{R}^{n_c}$, $\mathbf{U}  \in \mathbb{R}^{m \times n}$, and $\Gamma \in \mathbb{R}^{m}$ denote the joint position vector, mass/inertia matrix, sum of Coriolis/Centrifugal and gravity forces, constraint Jacobian, constraint force, selection matrix, and torque command, respectively. For the simple notations, let us consider $\mathbf{M}$, $b$, and $\mathbf{J}_{c}$ to be equal to $\mathbf{M}(q)$, $b(\dot{q}, q)$, and $\mathbf{J}_{c}(q)$. The constraints that we consider are $x_{c} = f_c(q) = \mathbf{c}$ where $f_c: \mathbb{R}^{n} \mapsto \mathrm{SE}(3)$ and $\mathbf{c}$ is a constant vector in $\mathrm{SE}(3)$, then, 
\begin{equation}
    \begin{split}
        \dot{x}_c &= \frac{\partial f_c}{\partial q} \dot{q} = \mathbf{J}_c \dot{q} =0, \\
        \ddot{x}_{c} & = \dot{\mathbf{J}}_{c} \dot{q} + \mathbf{J}_{c} \ddot{q} =0.
    \end{split}
\end{equation}
To incorporate these constraints in the equation of motion, the null-space projection matrix of the constraint Jacobian is defined as $\mathbf{N}_c = \mathbf{I} - \overline{\mathbf{J}}_c\mathbf{J}_c \in \mathbb{R}^{n \times n}$ where $\overline{\mathbf{J}}_c = \mathbf{M}^{-1}\mathbf{J}_c^{\top}(\mathbf{J}_c \mathbf{M}^{-1} \mathbf{J}_c^{\top})^{\dag}$. The constraint force $F_{c}$ can be obtained as follows:
\begin{equation}
    F_{c} = \overline{\mathbf{J}}_c^{\top} (\mathbf{U}^{\top} \Gamma_{a}  - b) + \mathbf{\Lambda}_{c} \dot{\mathbf{J}}_{c} \dot{q}
    \label{eq:contact_force}
\end{equation}
where $\mathbf{\Lambda}_c = (\mathbf{J}_c \mathbf{M}^{-1} \mathbf{J}_c^{\top})^{\dag}$. After substituting equation (\ref{eq:contact_force}) into (\ref{eq:dynamics}), we obtain the constrained dynamics equation of a robot as follows: 
\begin{equation}
  \mathbf{M} \ddot{q} + b_{c} = \mathbf{N}_c^{\top} \mathbf{U}^{\top} \Gamma \textrm{.}
  \label{eq:contact_eq}
\end{equation}
where $b_c =  \mathbf{N}_{c}^{\top} b + \mathbf{J}_{c}^{\top} \mathbf{\Lambda}_{c} \dot{\mathbf{J}}_{c} \dot{q}$. The dynamics equation in the constrained task space could be formulated by right multiplying the above equation by $\mathbf{J}_{1}\mathbf{M}^{-1}$, where $\mathbf{J}_{1}$ denotes the task Jacobian for $x_{1}$. The operational space dynamics then becomes
\begin{equation}
\begin{split}
\ddot{x}_{1} - \dot{\mathbf{J}}_{1|c}\dot{q} + \mathbf{J}_{1}\mathbf{M}^{-1} b_{c} = \mathbf{J}_{1|c} \mathbf{M}^{-1} \mathbf{U}^{\top} \Gamma
\label{eq:operational_dyn}
\end{split}
\end{equation}
where $\ddot{x}_{1}$ denotes the acceleration for the task $x_1$ in the constrained task space and $\mathbf{J}_{1|c} = \mathbf{J}_{1} \mathbf{N}_{c}$. Given $q$, $\dot{q}$ and the desired task acceleration, $\ddot{x}_1^{d}$, a constrained optimization problem is formulated to obtain the torque command as follows: 
\begin{equation} \label{cost}
    \begin{split}
        \min_{\Gamma} \quad  (\mathcal{M} \Gamma -\mathbf{b})^{\top} \mathbf{\Lambda}_{1|\mathrm{UN_c}}(\mathcal{M}\Gamma - \mathbf{b})
    \end{split}    
\end{equation}
where $\mathcal{M} =\mathbf{J}_{1|c} \mathbf{M}^{-1} \mathbf{U}^{\top}$, $\mathbf{b} = \ddot{x}_{1}^{d} - \dot{\mathbf{J}}_{1|c}\dot{q} + \mathbf{J}_{1|c}\mathbf{M}^{-1} b_c$, and $\mathbf{\Phi}^{-1} = \mathbf{U} \mathbf{M}^{-1} \left( \mathbf{U} \mathbf{N}_{c} \right)^{\top}$. The weighting matrix is computed as follows:
\begin{equation}
\begin{split}
    \bm{\Lambda}_{1|\mathrm{UN_c}}^{-1} =& \mathcal{M}\bm{\Phi}\mathcal{M}^{\top}\\
    =& \mathbf{J}_{1|c}\mathbf{M}^{-1}\mathbf{U}^{\top} \left(\mathbf{U}\mathbf{M}^{-1}\mathbf{N}_c^{\top}\mathbf{U}^{\top} \right)^{\dag} \mathbf{U}\mathbf{M}^{-1}\mathbf{J}_{1|c}^{\top}\\
    =&\mathbf{J}_{1}\overline{\mathbf{UN}}_{c} \mathbf{UN}_{c} \mathbf{M}^{-1}\mathbf{J}_{1}^{\top}
\end{split}
\end{equation}
where $\overline{\mathbf{UN}}_c \coloneqq \mathbf{M}^{-1}\mathbf{N}_c^{\top}\mathbf{U}^{\top} \left(\mathbf{U}\mathbf{N}_c\mathbf{M}^{-1}\mathbf{N}_c^{\top}\mathbf{U}^{\top} \right)^{\dag}$ and $\mathbf{N}_c = \mathbf{N}_c^{2}$ is an idempotent matrix. When $\overline{\mathbf{UN}}_{c} \mathbf{UN}_c = \mathbf{N}_c$, it is clear that $\bm{\Lambda}_{1|\mathrm{UN_c}}^{-1} = \mathbf{J}_1 \mathbf{N}_c\mathbf{M}^{-1}\mathbf{J}_{1}^{\top} = \bm{\Lambda}_{1|c}^{-1}$.

When there exists many solutions for $\Gamma$ to achieve $\mathcal{M}\Gamma = \mathbf{b}^{\star}$, WBC minimizes the weighted torque norm fulfilling the optimization
\begin{equation} \label{torque_opt}
    \begin{split}
        \min_{\Gamma}& \quad \Gamma^{\top} \bm{\Phi}^{-1} \Gamma \\
        \textrm{s.t.}& \quad \mathcal{M} \Gamma =  \mathbf{b}^{\star}.
    \end{split}
\end{equation}
When $\mathbf{b}^{\star} = \mathbf{b}$ and $ \overline{\mathbf{UN}}_c\mathbf{UN}_{c} = \mathbf{N}_{c}$, the optimal solution for the above optimization problem (\ref{torque_opt}) can be explicitly written as
\begin{equation}
\begin{split}
    \Gamma^{\star} &= \mathbf{\Phi}\mathcal{M}^{\top}(\mathcal{M}\mathbf{\Phi}\mathcal{M}^{\top})^{-1}\mathbf{b} = \bm{\Phi}\mathcal{M}^{-1}\bm{\Lambda}_{1|\mathrm{UN_c}}\mathbf{b} \\
    &= \overline{\mathbf{U}\mathbf{N}_c}^{\top}\mathbf{J}_{1|c}^{\top}\bm{\Lambda}_{1|\mathrm{UN_c}} \mathbf{b} = \overline{\mathbf{U}\mathbf{N}_c}^{\top}\mathbf{J}_{1|c}^{\top}\bm{\Lambda}_{1|c}\mathbf{b}
    \end{split}
\end{equation}
because $\mathbf{N}_{c}\mathbf{M}^{-1} = \mathbf{M}^{-1} \mathbf{N}_{c}^{\top}$. This control command for the task $x_1$ is identical to the WBC command proposed in \cite{sentis2010compliant}. What we've newly done above is to formalize the WBC controller as an optimization problem. One advantage of this optimization-form WBC is the ability to incorporate equality or inequality constraints embedded in the optimization problem. When we want to consider more constraints, it is possible to add explicit constrains directly. 

\subsection{Multiple Tasks with Hierarchies}
We consider multiple hierarchical tasks using WBC.    
\begin{definition}
Let us consider $n_t$ hierarchical tasks, $x_1, \: \cdots, \: x_{n_t}$. We can express a task hierarchy among the given tasks as $x_1 \gg \cdots \gg x_{n_t}$ where $x_{a} \gg x_{b}$ represents that $x_a$ has higher priority than $x_b$.
\end{definition}
The basic approach of WBC for multiple tasks is to employ lexicographical optimization. Given $n_t$ hierarchical tasks, the solution to the hierarchical WBC problem leads to the control command:
\begin{align} \label{wbc_torque}
    \Gamma^{\star} = & \overline{\mathbf{U}\mathbf{N}_c}^{\top} \mathbf{N}_c^{\top} \sum_{k=1}^{n_t}\Gamma_{k}  =\overline{\mathbf{U}\mathbf{N}_c}^{\top} \mathbf{N}_c^{\top} \sum_{k=1}^{n_t} \mathbf{J}_{\mathrm{prec}(k)}^{\top} \bm{\mathcal{F}}_{k}, \nonumber \\
    \bm{\mathcal{F}}_{k} =&   (\mathbf{J}_{\mathrm{prec}(k)|c}\mathbf{M}^{-1} \mathbf{U}^{\top} \overline{\mathbf{U}\mathbf{N}_c}^{\top}\mathbf{J}_{\mathrm{prec}(k)|c}^{\top})^{-1} \mathbf{b}_{k}\\
    =& \bm{\Lambda}_{\mathrm{prec}(k)|c} \mathbf{b}_{k} \quad(\textrm{when } \overline{\mathbf{UN}}_c \mathbf{UN}_{c} = \mathbf{N}_c), \nonumber\\
    \mathbf{b}_{k} =& \ddot{x}_{k}^{d} - \dot{\mathbf{J}}_{\mathrm{prec}(k)|c} \dot{q}  + \mathbf{J}_{\mathrm{prec}(k)|c} \mathbf{M}^{-1}b \nonumber
\end{align}
where $\mathbf{J}_{\mathrm{prec}(k)} = \mathbf{J}_{k}\mathbf{N}_{k-1}$, $\mathbf{N}_{k} = \mathbf{N}_{k-1} - \overline{\mathbf{J}}_{\mathrm{prec}(k)}\mathbf{J}_{\mathrm{prec}(k)}$, $\mathbf{N}_{0} = \mathbf{I}$, and $\mathbf{J}_{\mathrm{prec}(k)|c} = \mathbf{J}_{\mathrm{prec}(k)}\mathbf{N}_{c}$. We note that the tasks are controllable using actuated joints when $\overline{\mathbf{UN}}_c\mathbf{UN}_{c} = \mathbf{N}_{c}$, because
\begin{equation} \label{eq:const}
    \begin{split}
        \mathbf{M}\ddot{q} + \mathbf{N}_c^{\top} b + \mathbf{J}_c^{\top} \mathbf{\Lambda}_c \dot{\mathbf{J}}_c \dot{q} &= (\mathbf{U}\mathbf{N}_c)^{\top} \Gamma^{\star} \\
        &= \mathbf{N}_c^{\top} \sum_{k=1}^{n_t} \mathbf{J}_{\mathrm{prec}(k)}^{\top} \bm{\mathcal{F}}_k.
    \end{split}
\end{equation}
The task space dynamics for the $k$-th prioritized task $x_k$ are obtained by left-multiplying by $\mathbf{J}_{k}\mathbf{M}^{-1}$ as follows:
\begin{equation}
\begin{split}
    \ddot{x}_k  - \dot{\mathbf{J}}_{k|c} \dot{q} + \mathbf{J}_k\mathbf{M}^{-1} (\mathbf{N}_c^{\top}b + \mathbf{J}_c^{\top}\mathbf{\Lambda}_c \dot{\mathbf{J}}_{c} \dot{q}) \\
    = \mathbf{J}_k\mathbf{M}^{-1}\mathbf{N}_c^{\top}\sum_{j=1}^{k}\mathbf{J}_{\mathrm{prec}(j)}^{\top} \bm{\mathcal{F}}_j
\end{split}    
\end{equation}
where $\mathbf{J}_{k}\mathbf{M}^{-1}\mathbf{N}_c^{\top}\mathbf{J}_{\mathrm{prec}(j)}^{\top} \bm{\mathcal{F}}_j = 0$ for all $j>k$. Because 
\begin{equation*}
    \begin{split}
        \mathbf{J}_k \mathbf{M}^{-1} \mathbf{N}_{c}^{\top} \mathbf{J}_{\mathrm{prec}(j)}^{\top}\bm{\mathcal{F}}_{j} &= \mathbf{J}_k \mathbf{M}^{-1} \mathbf{N}_{c}^{\top} \mathbf{N}_{j-1}^{\top} \mathbf{J}_{j}^{\top} \bm{\mathcal{F}}_{j}\\
        &= \mathbf{J}_{k}\mathbf{N}_{j-1}\mathbf{N}_{c} \mathbf{M}^{-1}\mathbf{J}_{j}^{\top}\bm{\mathcal{F}}_j = 0
    \end{split}
\end{equation*}
where $\mathbf{J}_{k}\mathbf{N}_{j-1} = 0$ as shown in Appendix A in \cite{lee2012intermediate}. Based on the previous recursive null space projections and the above decoupled task space dynamics, the desired hierarchical tasks are effectively controlled in order of priority. Although this WBC projection-based method is straightforward, it does not allow to incorporate inequality constraints and it is only instantaneously optimal.

\section{The Proposed MPC}
\label{sec3}
We propose to replace WBC with MPC to execute multiple hierarchical tasks more efficiently. Before constructing an MPC, we specify the state space model of the robot dynamics from (\ref{eq:dynamics}) as follows:
\begin{equation} \label{state_model}
    \begin{split}
        \dot{\bm{x}}(t) &= f(\bm{x}(t)) + g(\bm{x}(t)) \bm{u}(t), \\
        f(\bm{x}(t)) &= \left[ \begin{array}{c} \dot{q} \\ -\mathbf{M}^{-1}b \end{array} \right], \\
        g(\bm{x}(t)) &= \left[ \begin{array}{cc}  \mathbf{0}_{n\times m} & \mathbf{0}_{n \times n_c} \\ \mathbf{M}^{-1}\mathbf{U}^{\top} & -\mathbf{M}^{-1}\mathbf{J}_{c}^{\top} \end{array} \right]
    \end{split}
\end{equation}
where $\bm{x} = [q^{\top},\:\dot{q}^{\top}]^{\top} \in \mathbb{R}^{n_{\bm{x}}}$, and $\bm{u} = [ \Gamma^{\top}, \: F_{c}^{\top} ]^{\top} \in \mathbb{R}^{n_{\bm{u}}}$. More specifically, the dimensions of the state and the input are $n_{\bm{x}} = 2n$ and $n_{\bm{u}} = m + n_c$. Given a finite-time horizon $[t_0, \: t_f]$, we formulate an optimal control problem as follows:
\begin{equation} \label{nonlinear_MPC}
    \begin{split}
        \min_{\bm{x}(.), \bm{u}(.)}&\quad \ell_f(\bm{x}(t_f)) + \int_{t_0}^{t_f} \ell(\bm{x}(t),\bm{u}(t)) dt \\
         \textrm{s.t.}&\quad  \dot{\bm{x}}(t) = f (\bm{x}(t))+ g(\bm{x}(t)) \bm{u}(t), \\
         &\quad h_i(\bm{x}(t),\bm{u}(t)) \leq 0, \\
         &\quad h_e(\bm{x}(t), \bm{u}(t)) = 0, \quad \bm{x}(t_0) = \bm{x}_0
    \end{split}
\end{equation}
where $h_i$ and $h_e$ are inequality and equality constraint functions, respectively. $\ell_f(.)$ and $\ell(.)$ are the cost at the terminal state $\bm{x}(t_f)$ and the running cost, respectively. In view of (\ref{cost}), the performance index for the WBC problem is equal to:
\begin{equation} \label{nonlinear_cost}
    \begin{split}
        &\ell(.) = \bm{u}^{\top} \mathbf{W}_{uu} \bm{u} - 2\mathbf{b}^{\top} \mathbf{W}_{bu}\bm{u} + \mathbf{b}^{\top} \bm{\Lambda}_{1|\mathrm{UN_c}}\mathbf{b}, \\
        &\ell_f(\bm{x}(t_f)) = \mathbf{b}^{\top}\bm{\Lambda}_{1|\mathrm{UN_c}} \mathbf{b}
    \end{split}
\end{equation}
where $\mathbf{W}_{uu} = \textrm{bdiag}(\mathcal{M}^{\top} \bm{\Lambda}_{1|\mathrm{UN_c}}\mathcal{M}, \mathbf{W}_c)$, $\mathbf{W}_{bu} = [\bm{\Lambda}_{1|\mathrm{UN_c}}\mathcal{M}, \: \mathbf{0}]$, and $\mathbf{W}_c \in \mathbb{S}_{+}^{n_c}$ denotes a weighting matrix for the constraint force. In addition, we choose the classical PD control law:
\begin{equation}
    \ddot{x}_1^{d}(t) = \mathbf{K}_p ( x_1^d(t) - x_1(t)) + \mathbf{K}_v (\dot{x}_1^d(t) - \dot{x}_1(t))    
\end{equation}
where $\mathbf{K}_p  = \textrm{diag}(K_{p_1}, \cdots, K_{p_{\dim(x_1)}})$ and $\mathbf{K}_v = \textrm{diag}(K_{v_1}, \cdots, K_{v_{\dim(x_1)}})$ are proportional and derivative gain matrices, respectively. We note that $\mathbf{W}_u$, $\mathbf{W}_{bu}$, $\bm{\Lambda}_{1|\mathrm{UN_c}}$, and $\mathbf{b}$ depend on the state $\bm{x}$. In addition, both the running cost $\ell$ and the final cost $\ell_f$ are nonlinear. The state space model of the system is also nonlinear. Therefore, we have formulated a nonlinear optimal control problem. The rest of this section explains the process of formulating this non-linear optimization as a convex MPC problem in the discrete time domain.

\subsection{QCQP to Control Hierarchical Tasks in the Discrete Domain}
\label{sec:MPC}
As a first step, we obtain a linearized state space model of the robotic systems in (\ref{state_model}). Consider the finite-time horizon $\mathbf{T}_{N} = [t_0, \: t_N]$. The time domain is normalized by using a dilation coefficient $\sigma = t_N - t_0$ and let $\tau = \sigma^{-1}(t- t_0) \in [0,\:1]$ for the unit interval. Then, we can convert the nonlinear dynamics of the robot as 
\begin{equation}\label{eq:newdyn}
    \dot{\bm{x}}_{\tau} = \frac{d \bm{x}_{\tau}}{dt} = \frac{d \bm{x}_{\tau}}{ \sigma d\tau} = f(\bm{x}_{\tau}) + g(\bm{x}_{\tau}) \bm{u}_{\tau}.
\end{equation}
Note that the dynamics in \eqref{eq:newdyn} are expressed in the normalized time domain. We now linearize these nonlinear dynamics given a reference trajectory $(\bm{x}_{\tau}^{d},\: \bm{u}_{\tau}^{d})$. By neglecting terms of order higher than $1$, this process produces the following approximated linear system
\begin{equation}
    d \bm{x}_{\tau} \approx (\mathbf{A}_{\tau}^{d} \bm{x}_{\tau} + \mathbf{B}_{\tau}^{d} \bm{u}_{\tau} + r_{\tau}^{d} )d\tau 
\end{equation}
where $r_{\tau}^{d} = \sigma\left[ f(\bm{x}_{\tau}^{d}) + g(\bm{x}_{\tau}^{d}) \bm{u}_{\tau}^{d}\right] - \mathbf{A}_{\tau}^{d} \bm{x}_{\tau}^d -  \mathbf{B}_{\tau}^{d} \bm{u}_{\tau}^{d}$, $\mathbf{A}_\tau^{d} = \sigma \nabla_{\bm{x}} (f(\bm{x})+g(\bm{x})\bm{u})|_{(\bm{x}_{\tau}^{d}, \bm{u}_{\tau}^{d})}$, and $\mathbf{B}_{\tau}^{d} =  \sigma \nabla_{\bm{u}} (f(\bm{x})+g(\bm{x})\bm{u})|_{(\bm{x}_{\tau}^{d}, \bm{u}_{\tau}^{d})}=\sigma g(\bm{x}_{\tau}^{d})$. A simple method to obtain the discrete-time state space model is to integrate the above differential equation: 
\begin{equation}
    \int_{\tau_{i}}^{\tau_{i} +\Delta \tau } d\bm{x}_{\tau} = \int_{\tau_{i}}^{\tau_{i} +\Delta \tau } \left(\mathbf{A}_{\tau}^{d} \bm{x}_{\tau} + \mathbf{B}_{\tau}^{d} \bm{u}_{\tau} + r_{\tau}^{d} \right) d\tau
\end{equation}
from which we obtain the following discrete-time state space model:
\begin{equation} \label{discrete_time}
    \bm{x}_{i+1} = \mathbf{A}_{i}^{d} \bm{x}_{i} + \mathbf{B}_{i}^{d} \bm{u}_{i} + r_{i}^{d}
\end{equation}
where $\mathbf{A}_{i}^{d} = \mathbf{A}_{\tau_i}^{d} \Delta \tau +\mathbf{I}$, $\mathbf{B}_{i}^{d} = \mathbf{B}_{\tau_i}^{d}\Delta \tau$, and $r_{i}^{d}= r_{\tau_i}^{d}\Delta \tau$. The concatenated state vector and control input are defined as
\begin{equation}
    \begin{split}
        \bm{\mathcal{X}}_{i} &= [\bm{x}_0^{\top},\: \bm{x}_{1}^{\top}, \: \cdots, \bm{x}_{i}^{\top}]^{\top} \in \mathbb{R}^{(i+1)n_x}, \\ 
        \bm{\mathcal{U}}_{i} &= [\bm{u}_0^{\top}, \: \bm{u}_{1}^{\top}, \: \cdots, \bm{u}_{i}^{\top}]^{\top} \in \mathbb{R}^{(i+1)n_u}.
    \end{split}
\end{equation}
Using these vectors, we can re-write the state space model as 
\begin{equation}\label{discrete_modi}
\begin{split}
    \bm{x}_{i} &= \bm{\mathcal{A}}_{i}\bm{x}_0 + \bm{\mathcal{B}}_{i}\bm{\mathcal{U}}_{i-1} + \bm{\mathcal{R}}_{i} \mathbf{1}_{i},\\
    \bm{\mathcal{B}}_{i} &= [\mathcal{B}_{i-1|0}, \: \cdots, \: \mathcal{B}_{i-1|i-2}, \: \mathcal{B}_{i-1|i-1}], \\
    \bm{\mathcal{R}}_{i} &= [\mathcal{R}_{i-1|0}, \: \cdots, \: \mathcal{R}_{i-1|i-2}, \: 
\mathcal{R}_{i-1|i-1}]
\end{split}
\end{equation}
where $\bm{\mathcal{A}}_{i} = \prod_{j=0}^{i} \mathbf{A}_{i-j}^{d}$ when $i\geq 1$ and $\bm{\mathcal{A}}_0 = \mathbf{I}$. In addition, $\mathcal{B}_{i|i-\beta} = (\prod_{j=0}^{\beta-1} \mathbf{A}_{i-j}^{d}) \mathbf{B}_{i-\beta}^{d}$, and $\mathcal{R}_{i|i-\beta} = (\prod_{j=0}^{\beta-1} \mathbf{A}_{i-j}^{d}) r_{i-\beta}^{d}$ when $\beta \geq 1$. Otherwise, when $\beta =0$, $\mathcal{B}_{i|i} = \mathbf{B}_{i}^{d}$ and $\mathcal{R}_{i|i} = r_{i}^{d}$, respectively. By concatenating the equation (\ref{discrete_modi}) for all $i \in \{0, \: \cdots, \: N\}$, the state equation can be written as follows:
\begin{equation}
    \bm{\mathcal{X}}_{N} = \bm{\mathcal{A}}^{d} \bm{x}_0 + \bm{\mathcal{B}}^{d} \bm{\mathcal{U}}_{N-1} + \bm{\mathcal{R}}_{N}^{d} \mathbf{1}_{n_{\bm{\mathcal{X}}}}
\end{equation}
where $n_{\bm{\mathcal{X}}} = \dim(\bm{\mathcal{X}}) = (N+1)n_{\bm{x}}$. Also $\bm{\mathcal{A}}^{d}$, $\bm{\mathcal{B}}^{d}$, and $\bm{\mathcal{R}}_{N}^{d}$ are formed by stacking the terms from $i=0$ to $i=N$ in (\ref{discrete_modi}). 

\begin{definition} \label{def1}
Consider $n_t$ hierarchical tasks, $x_1 \gg x_2 \gg \cdots \gg x_{n_t}$. Let the position trajectories, $x_{k}^{d}(t)$, be given. We can also express the hierarchy in terms of the resulting task tracking errors over a finite-time horizon $[t_0, t_f]$ as follows:
\begin{equation*}
    \begin{split}
        \| \mathbf{e}_{1}(t) \|^{2}  + \epsilon_{1} \leq \cdots \leq  \| \mathbf{e}_{n_t}(t) \|^{2} + \epsilon_{n_t}
    \end{split}
\end{equation*}
where $\mathbf{e}_{k}(t) = x_{k}^{d}(t) - x_{k}(t)$ for all $t\in[t_0, t_f]$. In addition, $\epsilon_{k} \geq 0$ where $\epsilon_{k-1} \leq \epsilon_{k}$, $k \in \{1, \cdots, n_t \}$, and $\epsilon_{0} = 0$.
\end{definition}

Let $q_i = q(t_i)$ and $\dot{q}_i = \dot{q}(t_i)$ where $t_i \in [t_0, \: t_N]_d$. We can specify and approximate the constraint $\| \mathbf{e}_k(q_i) \| + \epsilon_{k} \leq \| \mathbf{e}_{k+1} (q_i) \| + \epsilon_{k+1}$ where $\mathbf{e}_k(q_i) = x_k^{d}(t_i) - f_{t_k}(q_i)$ with $f_{t_k}:\mathbb{R}^{n} \mapsto \mathbb{R}^{\dim(x_k)}$ being a continuous function for the $k$-th task $x_k$ as follows:
\begin{align}
    &\| \mathbf{e}_k(q_i) \|^{2} - \| \mathbf{e}_{k+1}(q_i) \|^{2} + \epsilon_{k} - \epsilon_{k+1} \\
    &\approx (q_{i}^d - q_{i})^{\top} \left( \mathbf{J}_{k_{i}^d}^{\top}\mathbf{J}_{k_{i}^{d}} - \mathbf{J}_{{k+1}_{i}^d}^{\top}\mathbf{J}_{{k+1}_{i}^{d}} \right) (q_{i}^{d} - q_{i}) + \epsilon_{k(k+1)} \nonumber \\
    &= q_{i}^{\top} \mathcal{J}_{k_{i}^d} q - 2 q_{i}^{\top} \mathcal{J}_{k_{i}^d} q_{i}^{d} + q_{i}^{d\top} \mathcal{J}_{k_{i}^{d}} q_{i}^d + \epsilon_{k(k+1)} \leq 0 \nonumber
\end{align}    
where $\mathcal{J}_{k_{i}^d} =\mathbf{J}_{k_{i}^{d}}^{\top}\mathbf{J}_{k_{i}^{d}} - \mathbf{J}_{{k+1}_{i}^d}^{\top}\mathbf{J}_{{k+1}_{i}^{d}}$, $\mathbf{J}_{{k}_{i}^{d}} = \frac{\partial f_{t_k}}{\partial q}(q_{i}^d)$, and $\epsilon_{k(k+1)} = \epsilon_{k} - \epsilon_{k+1}$. Now, the above approximated constraints are convex quadratic functions. The concatenated form of the above equations is as follows:
\begin{equation}\label{quad_const}
    \bm{\mathcal{X}}_{N}^{\top}\bm{\mathcal{J}}_{k_{i}}^d \bm{\mathcal{X}}_{N} + \bm{\mathcal{Z}}_{k_i}^{d}\bm{\mathcal{X}}_{N} + \bm{\mathcal{E}}_{k_i}^d \leq 0
\end{equation}
where $\bm{\mathcal{J}}_{k_i}^{d} = \textrm{bdiag}(\mathbf{0},\cdots,  \widehat{\mathcal{J}}_{k_{i}}, \cdots, \mathbf{0}) \in \mathbb{R}^{n_{\bm{\mathcal{X}}} \times n_{\bm{\mathcal{X}}}}$, $\bm{\mathcal{Z}}_{k_i}^{d} = [ \mathbf{0},\cdots, \widehat{\mathcal{Z}}_{k_i}^{\top}, \cdots, \mathbf{0}] \in \mathbb{R}^{1\times n_{\bm{\mathcal{X}}}}$ , and $\bm{\mathcal{E}}_{k_i}^{d} = q_{k_i}^{d\top} \mathcal{J}_{k_i^d}q_{k_i}^{d} + \epsilon_{k(k+1)}$. Each sub-matrix is specified as follows: 
\begin{equation}
    \begin{split}
       \widehat{\mathcal{J}}_{k_i} &= \left[ \begin{array}{cc}\mathcal{J}_{{k}_{i}^d} & \mathbf{0}\\ \mathbf{0} & \mathbf{0} \end{array}\right] \in \mathbb{R}^{n_x\times n_x},\\ 
       \widehat{\mathcal{Z}}_{k_i} &= \left[ (-2q_{k_i}^{d\top}\mathcal{J}_{k_i^d})^{\top}, \: \mathbf{0} \right]^{\top} \in \mathbb{R}^{n_x}.
    \end{split}
\end{equation}
We consider the quadratic constraints expressed by (\ref{quad_const}) for all $k\in \{1, \cdots, n_t\}$ and $i\in \{1, \cdots, N\}$ then simply express the entire quadratic inequality constraint as $\bm{\mathcal{G}}(\bm{\mathcal{X}}_N) \leq 0$.
The case $\epsilon_{(k-1)k}=0$ is called as \textit{weak hierarchy}, which means to allow that the tracking error of the higher prioritized task can be equal to that of the lower prioritized task. The case, $\epsilon_{(k-1)k} < 0$, the error norm for the $(k-1)$-th task $\|\mathbf{e}_{k-1}\|$ is strictly smaller than the $k$-th task error $\|\mathbf{e}_{k}\|$, which is called a \textit{strong hierarchy}.  

Thirdly, we construct a convex (quadratic) approximation of the nonlinear performance index in (\ref{nonlinear_cost}) to make the problem tractable. We aim to solve the nonlinear optimal control problem in (\ref{nonlinear_MPC}), which we call $\mathbf{P}_{QCQP}( \bm{x}_0, \mathbf{T}_{N})$ as follows: 
\begin{equation} \label{eq:MPC}
    \begin{split}
        \min_{\bm{\mathcal{X}}_N, \bm{\mathcal{U}}_{N-1}} &\quad  \bm{\mathcal{L}}(\bm{\mathcal{X}}_{N}, \bm{\mathcal{U}}_{N-1})\\
         \textrm{s.t.}&\quad  \bm{\mathcal{X}}_{N} = \bm{\mathcal{A}}^{d} \bm{x}_0 + \bm{\mathcal{B}}^{d} \bm{\mathcal{U}}_{N-1} + \bm{\mathcal{R}}_{N}^{d} \mathbf{1}_{n_{\bm{\mathcal{X}}}}, \\
         &\quad \bm{\mathcal{G}}( \bm{\mathcal{X}}_N) \leq 0,\\
         &\quad \bm{\mathcal{H}}( \mathbf{\bm{\mathcal{X}}}_{N}) = 0,\\
         &\quad \bm{x}(t_0) = \bm{x}_0
    \end{split}
\end{equation}
and 
\begin{equation}
\begin{split}
        \bm{\mathcal{L}}(\bm{\mathcal{X}}_N, \bm{\mathcal{U}}_{N-1}) =& \bm{\mathcal{X}}_{N}^{\top} \bm{\mathcal{W}}_{xx}\bm{\mathcal{X}}_{N}  + \bm{\mathcal{W}}_{x} \bm{\mathcal{X}}_{N}   \\
        &+  \bm{\mathcal{U}}_{N-1}^{\top} \bm{\mathcal{W}}_{uu}\bm{\mathcal{U}}_{N-1} + \bm{\mathcal{W}}_{u} \bm{\mathcal{U}}_{N-1}
\end{split}
\end{equation}
where $\bm{\mathcal{H}}: \mathbb{R}^{n_{\bm{\mathcal{X}}}} \mapsto \mathbb{R}^{(N+1)n_c}$ is the linearized constraint function in terms of the stacked state vector $\bm{\mathcal{X}}_{N}$. In detail, the kinematic constraint is approximated as follows:
\begin{equation}
    f_c(q_i) \approx f_c(q_i^{d}) + \mathbf{J}_{c}(q_i^{d})(q_i^{d} - q_i) = \mathbf{c}
\end{equation}
where $q_i^{d}$ denote the nominal joint position in the $i$-th discrete time step. In turn we can express the above equality constraint in terms of the state.
\begin{equation}
    \left[\begin{array}{cc}
         -\mathbf{J}_c(q_i^d) & \mathbf{0}  
    \end{array} \right] \bm{x}_{i} + \left[\begin{array}{c}
         \mathbf{J}_c(q_i^d)q_{i}^{d} + f_c(q_i^{d}) - \mathbf{c}  
    \end{array} \right] = \mathbf{0}
\end{equation}
We consider this linear equality constraint for all $i \in \{1, \cdots, N\}$ in the convex optimization by concatenating in an appropriate form. In addition, $\bm{\mathcal{W}}_{xx}$, $\bm{\mathcal{W}}_{x}$, $\bm{\mathcal{W}}_{uu}$, and $\bm{\mathcal{W}}_{u}$ represent the weighting matrices, respectively. We utilize the nominal trajectories to shape the quadratic cost by assuming there exists feedback control gains $\mathbf{K}_{p}$ and $\mathbf{K}_{v}$ for $\check{\mathbf{b}}$:
\begin{equation*}
    \check{\mathbf{b}}_{i} \approx \underbrace{\left[\begin{array}{cc} -\check{\mathbf{K}}_p \check{\mathbf{J}}_{i^{d}} &  - \check{\mathbf{K}}_v \check{\mathbf{J}}_{i^{d}} \end{array} \right]}_{\mathbf{C}_{i}^d} \bm{x}_{i} + \underbrace{\left[ \begin{array}{c} \check{\mathbf{K}}_{p} \check{\mathbf{J}}_{i^d} q_{i}^{d} + \check{\mathbf{K}}_v \psi_{i^{d}} \end{array} \right]}_{\mathbf{c}_{i}^{d}}
\end{equation*}
where $\psi_{k_i^d} = \left[ \dot{x}_{1}^{d\top}(t_i), \cdots,  \dot{x}_{n_t}^{d\top}(t_i) \right]^{\top}$, $\check{\mathbf{K}}_{p} = \textrm{bdiag}(\mathbf{K}_{p_1},$ $\cdots, \mathbf{K}_{p_{n_t}})$, $\check{\mathbf{K}}_{v} = \textrm{bdiag}(\mathbf{K}_{v_1}, \cdots, \mathbf{K}_{v_{n_t}})$, and $\check{\mathbf{J}}_{i^d}$ is the stack of the Jacobians for all tasks such as 
\begin{equation}
    \check{\mathbf{J}}_{i^d} = \left[\begin{array}{c}
           \mathbf{J}_{1_i^d}\\
           \vdots \\
           \mathbf{J}_{{n_t}_i^{d}}
    \end{array} \right] \in \mathbb{R}^{\sum_{i=1}^{n_t} \dim (x_i)\times n}.
\end{equation}
Then, the running and final costs can be approximated as 
\begin{equation*}
    \begin{split}
        &\widetilde{\ell}(t_i) = \bm{u}_{i}^{\top} \mathbf{W}_{uu_i}^{d} \bm{u}_{i} + \bm{x}_{i}^{\top} \mathbf{W}_{xx_i}^{d} \bm{x}_{i} + \mathbf{W}_{u_i}^{d} \bm{u}_{i} + \mathbf{W}_{x_i}^{d} \bm{x}_{i} + \mathcal{C}_{1},\\
        &\widetilde{\ell_f}(\bm{x}(t_{N})) = \bm{x}_{N}^{\top}  \mathbf{W}_{xx_N}^{d} \bm{x}_N + \mathbf{W}_{x_N}^{d} \bm{x}_{N} +\mathcal{C}_2
    \end{split}
\end{equation*}
where 
\begin{equation}
    \begin{split}
        \mathbf{W}_{uu_i}^d =& \mathbf{W}_{uu}|_{(\bm{x}_i^d)}, \\
        \mathbf{W}_{xx_i}^{d} =&\mathbf{C}_{i}^{d\top} \bm{\Lambda}_{\check{i}|\mathrm{UN_c}} \mathbf{C}_{i}^{d}|_{(\bm{x}_i^d)},\\
        \mathbf{W}_{u_i}^{d} =& -2 \check{\mathbf{b}}_{i}^{\top} \mathbf{W}_{bu}|_{(\bm{x}_i^d)},\\
        \mathbf{W}_{x_i}^{d} =& 2 \mathbf{c}_{i}^{d\top} \bm{\Lambda}_{\check{i}|\mathrm{UN_c}} \mathbf{C}_{i}^{d}|_{(\bm{x}_i^d)}
    \end{split}
\end{equation}
$\bm{\Lambda}_{\check{i}|\mathrm{UN_c}}$ is the task inertia matrix computed by using the stacked Jacobian $\check{\mathbf{J}}_{i^d}$. $\mathcal{C}_{1}$ and $\mathcal{C}_2$ is the sum of the remaining terms in the running and terminal costs, which are dropped in the quadratic approximation of the latter. The approximated running cost can be stacked for the augmented vectors $\bm{\mathcal{X}}_{N}$ and $\bm{\mathcal{U}}_{N-1}$ such that 
\begin{equation}
    \begin{split}
        \bm{\mathcal{W}}_{uu} =& \textrm{bdiag}(\mathbf{W}_{uu_0}^{d}, \cdots, \mathbf{W}_{uu_{N-1}}^{d}),\\
        \bm{\mathcal{W}}_{xx} =& \textrm{bdiag}(\mathbf{W}_{xx_0}^{d}, \cdots, \mathbf{W}_{xx_{N}}^{d}),\\
        \bm{\mathcal{W}}_{u} =& [ \mathbf{W}_{u_0}^{d}, \cdots, \mathbf{W}_{u_{N-1}}^{d} ],\\
        \bm{\mathcal{W}}_{x} =& [ \mathbf{W}_{x_0}^{d}, \cdots, \mathbf{W}_{x_{N}}^{d} ].
    \end{split}
\end{equation}
Finally, the formulated MPC problem in (\ref{eq:MPC}) becomes a QCQP by approximating the performance index, the system dynamics, and the constraints along the nominal trajectories. The detailed process is described in Algorithm 1.

\subsection{Nominal Trajectories from IK and ID}
We assume trajectories for the hierarchical tasks, $x_1 \gg x_2 \gg \cdots \gg x_{n_t}$, are given over a  finite-time horizon $x_k^d(t)$ where $k\in \{1, \cdots, n_t\}$ and $t\in [t_0, t_f]$. We need to convert these task trajectories into state space reference trajectories to be employed in our MPC. In this section, we obtain nominal trajectories in joint position and velocity space by solving the inverse kinematics problem. Let the initial state be given as $\bm{x}_0 = [q_0^{\top}, \: \dot{q}_0^{\top}]^{\top}$. We can recursively compute the nominal trajectories with respect to the desired task specifications. Let us consider the discretized time domain as described in Section \ref{sec:MPC}. We start from $x_{k_{i+1}}^{d} -x_{k_{i}}^{d} = \mathbf{J}_{k_{i}^{d}}(q_{i+1}^{d} - q_{i}^{d})$. We can update the desired joint velocity for $n_t$ hierarchical tasks using the null space projection method proposed in \cite{lee2012intermediate}: 
\begin{equation}
\begin{split}
    \bm{\mathcal{Q}}_{i} &= \mathbf{J}_{1_i^d}^{\dag}(x_{1_{i+1}}^{d} - x_{1_i}^d) +  \sum_{k=2}^{n_t} \Delta \bm{q}_{k_i},\\
    \Delta \bm{q}_{k_i} &= (\mathbf{J}_{k_{i}^{d}}\mathbf{P}_{{k-1}_{i}^{d}})^{\dag}\left( x_{k_{i+1}}^{d} - x_{k_{i}}^{d} - \mathbf{J}_{k_{i}^{d}} \Delta \mathbf{q}_{k-1_{i}} \right), \\
    \mathbf{P}_{k_i^{d}} &= \mathbf{P}_{{k-1_i^{d}}} - (\mathbf{J}_{k_i^{d}}\mathbf{P}_{k-1_i^d})^{\dag} (\mathbf{J}_{k_i^d}\mathbf{P}_{k-1_i^d})
\end{split}
\end{equation}
where $\mathbf{P}_{0_i^d} = \mathbf{I}$, and $\Delta \bm{q}_{0} = \mathbf{0}$. The $(i+1)$-th desired value for the state $\bm{x}_i^{d} = [q_{i}^{\top}, \: \dot{q}_i^{\top}]^{\top}$ is obtained as follows:
\begin{equation}
\begin{split}
    q_{i+1}^{d} &= q_{i}^{d} + \bm{\mathcal{Q}}_{i},\\
    \dot{q}_{i+1}^{d} &= (q_{i+1}^{d}- q_{i}^{d}) \sigma N^{-1}  
\end{split}
\end{equation}
where $i \in \{ 0, \: \cdots, \: N-1\} $, $q_{0}^{d} = q_{0}$, and $\dot{q}_0^{d} =\dot{q}_0$. Given the desired state trajectories $\bm{x}_{i}^d$, WBC can be utilized to obtain the instantaneous input reference, $\bm{u}_{i}^{d}$ to control the hierarchical tasks as described in (\ref{wbc_torque}). These nominal trajectories for the state and input are utilized to compute the matrices $\bm{\mathcal{A}}^d$, $\bm{\mathcal{B}}^{d}$, $\bm{\mathcal{R}}_{N}^{d}$, $\bm{\mathcal{W}}_{xx}$, $\bm{\mathcal{W}}_{x}$, $\bm{\mathcal{W}}_{uu}$, and $\bm{\mathcal{W}}_{u}$ in (\ref{eq:MPC}).

\subsection{The proposed convex MPC}
Based on the formulated QCQP, we construct a convex MPC problem considering an $N_p$ prediction horizon, $\mathbf{T}_{p|s}=[t_{sN_e}, \: t_{sN_e+N_p}]_{d}$ and an $N_e$ execution horizon, $\mathbf{T}_{e|s}=[t_{sN_e}, \: t_{(s+1)N_e}]_{d}$ where $s\in \{0, \: \cdots , \: N_{e}^{-1}N -1\}$. Our MPC consists of an iterative process solving the formulated QCQP over different prediction horizons as shown in Algorithm \ref{algorithm1}. The output of this algorithm consists of the entire state trajectory $\bm{\mathcal{X}}^{\star}$ and the corresponding control input $\bm{\mathcal{U}}^{\star}$. 

\begin{algorithm}[h]
 \KwData{$\bm{x}_0$, $\mathbf{T}_N=[t_0, t_f]_d$, $x_{k}^d(t)$ where $k\in \{ 1, \cdots, n_t\}$, $t\in \mathbf{T}_N$, $x_{1} \gg \cdots \gg x_{n_t}$ }
 \KwResult{$\bm{\mathcal{X}}^{\star}$ and $\bm{\mathcal{U}}^{\star}$  }
 $\bm{x}_{(0:N)}^{d} \leftarrow $ IK w.r.t. $x_{k}^{d}(t_{i})$ for all $k\in \{ 1, \cdots, n_t\}$\; 
 $\bm{u}_{(0:N-1)}^{d} \leftarrow $ ID w.r.t. $x_{k}^{d}(t_{i})$ for all $k\in \{ 1, \cdots, n_t\}$\; 
 $\widetilde{\bm{x}} \leftarrow \bm{x}_0$, $\bm{\mathcal{X}}^{\star} \leftarrow \emptyset$, $\bm{\mathcal{U}}^{\star} \leftarrow \emptyset$  \;
	\For{$s\leftarrow 0$ \KwTo $N_e^{-1}N -1$}{
        $(\bm{\mathcal{X}}_{N_p}^{*}, \bm{\mathcal{U}}_{N_p-1}^{*}) \leftarrow \mathbf{P}_{QPQC}(\widetilde{\bm{x}}, \mathbf{T}_{p|s})$ in (\ref{eq:MPC}) \;
        $\bm{\mathcal{X}}^{\star} \leftarrow [\bm{\mathcal{X}}^{\star \top}, \: \bm{\mathcal{X}}_{N_e+1}^{* \top} ]^{\top}$, $\bm{\mathcal{U}}^{\star} \leftarrow [\bm{\mathcal{U}}^{\star \top}, \: \bm{\mathcal{U}}_{N_e}^{* \top} ]^{\top}$ \;
        $\widetilde{\bm{x}} \leftarrow \bm{x}_{N_e+1}^{*}$ from $\bm{\mathcal{X}}_{N_p}^{*}$ \;
 	}
\caption{Algorithm for the proposed MPC}
\label{algorithm1}
\end{algorithm}

\begin{figure}
\centering
\includegraphics[width=0.8\linewidth]{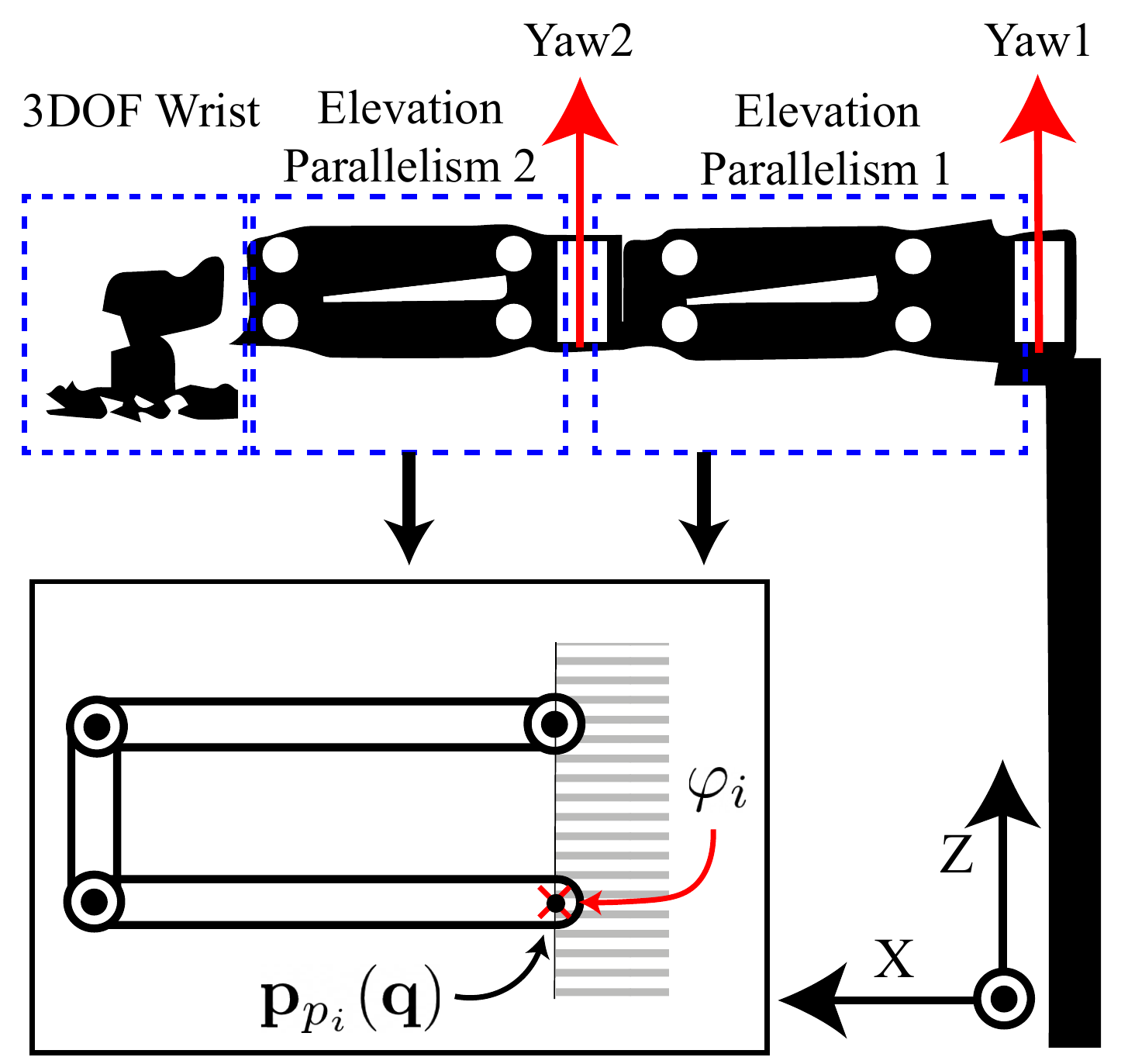}
\caption{Scorpio model and parallelograms: $\mathbf{P}_{p_i}(q)$ and $\varphi_i$ denote the pivoting end-part and constrained position of the $i$-th parallelogram. }
\label{Fig1}
\end{figure}

\section{Numerical Simulation}
\label{sec4}
In this section, we validate the proposed convex MPC-based approach by using \textit{Scorpio}, which is a unique type of robotic manipulator. We briefly introduce the manipulator including mechanical parallelisms and demonstrate the numerical simulations for the proposed convex MPC-based approach. We compare the results of the proposed method with those obtaining by applying WBC to show its efficiency. The simulation is implemented on a laptop with MATLAB\footnote{The MathWorks Inc., MATLAB, Version R2019b, Natick, MA (2018) } and we obtain analytic expressions of the terms in the state equation by using Mathematica\footnote{Wolfram Research, Inc., Mathematica, Version 12.0, Champaign, IL (2019).} and FROST \cite{Hereid2017FROST}. 

\subsection{Underactuated and Constrained Robotic Manipulator}
\textit{Scorpio} is a unique robotic manipulator that is designed to efficiently handle heavy objects using low power. In particular, two mechanical parallelograms compensate for the gravitational force of the robot's load, enhancing its lifting capabilities. However, many complicated problems, i.e., passive joints and constraints, arise due to the use of the unique mechanical structures as shown in Fig \ref{Fig1}. More specifically, the robot has $11$ DOFs and $4$ of them are passive joints describing the parallelograms' motions. For each parallelogram, the $y$ position in the body frame is not controllable because of the type of mechanical structure. Therefore, the constraint Jacobian for each parallelogram is computed as follows:
\begin{equation}
    \begin{split}
        \mathbf{J}_{c}(q) = \left[\begin{array}{c} \mathbf{J}_{p_i}^{x}(q) \\
        \mathbf{J}_{p_i}^{z}(q) \end{array} \right], \quad  \mathbf{J}_{p_i}(q) = \left[ \begin{array}{c} \mathbf{J}_{p_i}^{x}(q) \\
        \mathbf{J}_{p_i}^{y}(q) \\
        \mathbf{J}_{p_i}^{z}(q) \end{array} \right]
    \end{split}
\end{equation}
where $\mathbf{J}_{p_i}(q) = \frac{\partial \mathbf{P}_{p_i}}{\partial q}(q) \in \mathbb{R}^{3\times 11}$. The dimension of the constrained dynamics described using the null space matrix $\mathbf{N}_c$ becomes $6$ which is identical to the number of active joints. 

\begin{figure}
\centering
\includegraphics[width=\linewidth]{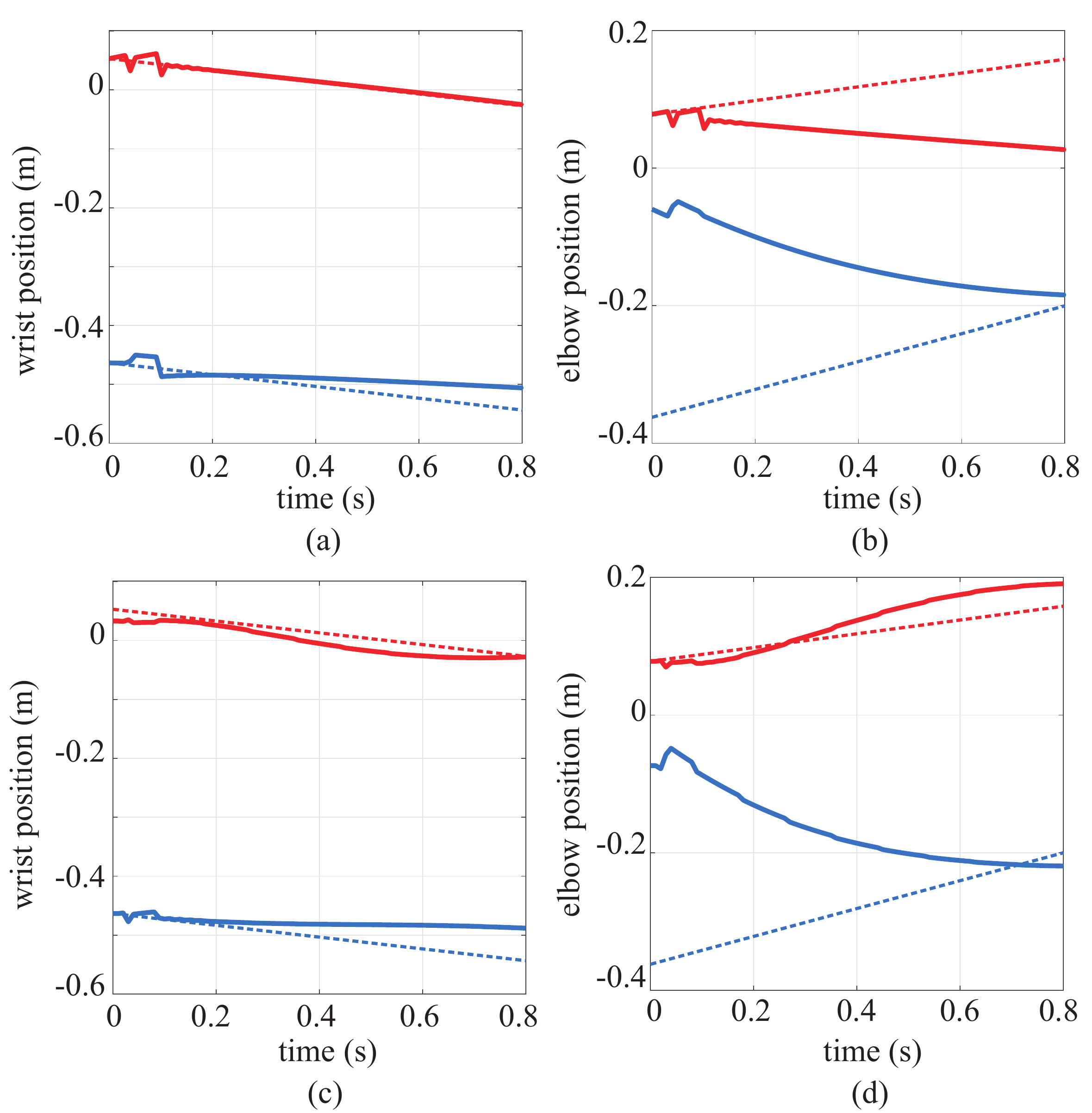}
\caption{Elbow and wrist positions in Cartesian space: (a) wrist position controlled by WBC, (b) elbow position controlled by WBC, (c) wrist position controlled by MPC, (d) elbow position controlled by MPC. Dotted lines represent the desired trajectories and solid lines indicate the results achieved using WBC or MPC. Red and blue lines in (a) and (c) represent the data in the $x$ and $y$ directions. In (b) and (d), $x$ and $z$ positions are represented by red and blue lines, respectively.}
\label{Fig2}
\end{figure}

\subsection{Execution of Two Hierarchical Tasks}
In this numerical simulation, we define two hierarchical tasks: control of the elbow position in the x and z directions, $x_e \in \mathbb{R}^{2}$, and control of the wrist position in the x and y directions, $x_w \in \mathbb{R}^{2}$ where $x_w \gg x_e$. For the sake of simplicity, we reduce the dimension of the state by making the $3$ wrist joints completely rigid because the wrist joints do not affect the defined tasks. In this simulation, we set $\mathbf{K}_p = \textrm{diag}(40,\: 40)$ and $\mathbf{K}_v = \textrm{diag}(2,\: 2)$. The full time horizon is defined as $\mathbf{T}_N = [ t_0, \: t_f] = [0, \: 0.8]$ with $0.01 \:s$ time increments, which means $N=80$. We set the prediction and execution steps as $N_p=10$ and $N_e =3$, respectively. The initial configuration of the robot is $[ -90^{\circ}, \: 0^{\circ},\: 0^{\circ}, \:0^{\circ}, \: -90^{\circ}]$. We set the desired trajectories of both tasks using a linear interpolation between the initial and final positions. More specifically, we consider $x_e(t_0) = [0.0780, \: -0.3622]$, $x_e(t_f) = [0.1780, \: -0.1597]$, $x_w(t_0) = [0.0531 ,\: -0.4634]$, and $x_w(t_f) = [ -0.0469, \: -0.5634]$.

\begin{figure}
\centering
\includegraphics[width=\linewidth]{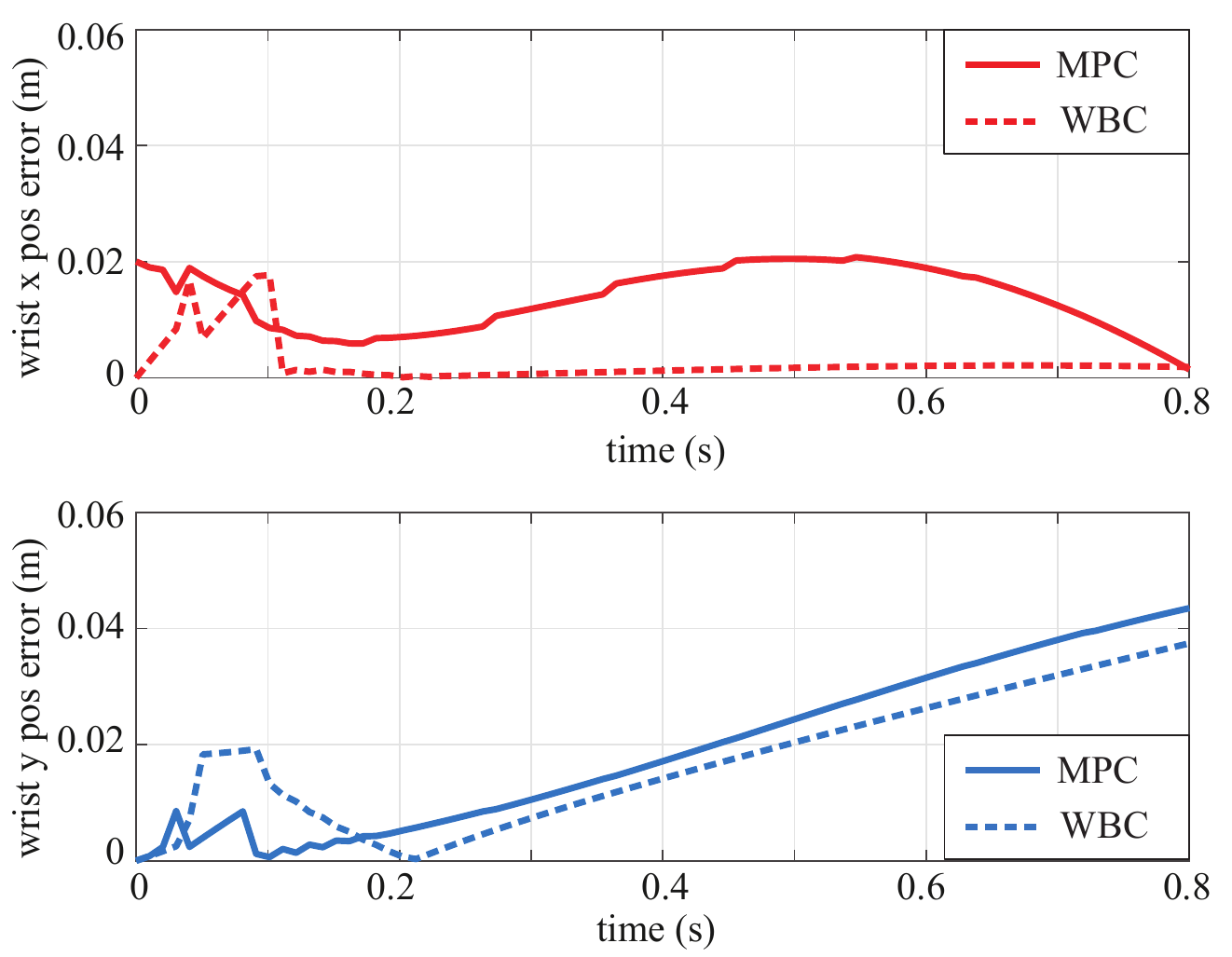}
\caption{Error of the wrist positioning task: the upper graph shows the wrist positioning error in the $x$ direction and the lower graph shows the positioning error in the $y$ direction.}
\label{Fig3}
\end{figure}

\subsection{Comparison with WBC}
In this section, we compare the simulation results controlled by the proposed MPC controller with those executed by WBC as described in Section \ref{sec:2}. Fig. \ref{Fig2} shows the simulation results implemented by both WBC and the proposed MPC. Firstly, WBC instantaneously minimizes the positioning error by considering the task hierarchy shown (a) and (b) in Fig. \ref{Fig2}. WBC minimizes the higher prioritized task $x_w$ error. Sequentially, the lower prioritized task $x_e$ is controlled by keeping the optimized task error for $x_w$. On the other hand, the proposed MPC considers the finite-time prediction horizon and we do not have cascaded optimization structures. For these reasons, the wrist positioning task $x_w$ has a little bit larger errors than those by WBC as shown (a) and (c) in Fig. \ref{Fig2}. However, the proposed MPC-based approach reduces the errors of the elbow positioning task which has lower hierarchy as shown in (b) and (d) subfigures of Fig. \ref{Fig2}. 

\begin{figure}
\centering
\includegraphics[width=\linewidth]{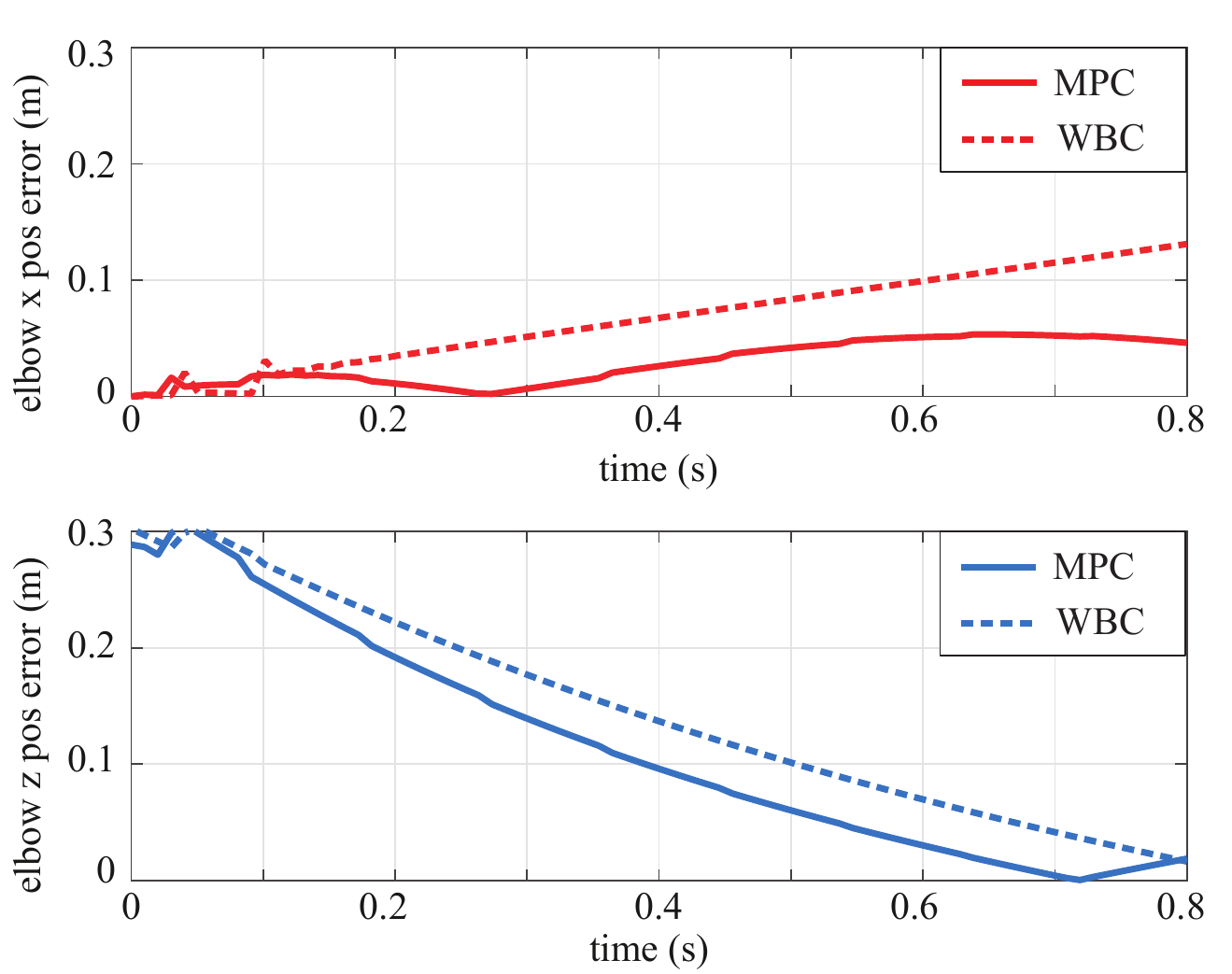}
\caption{Error of the elbow positioning task: the upper and lower graphs show the position errors of the elbow positioning task in the $x$ and $z$ directions, respectively.  }
\label{Fig4}
\end{figure}

\begin{figure}
\centering
\includegraphics[width=\linewidth]{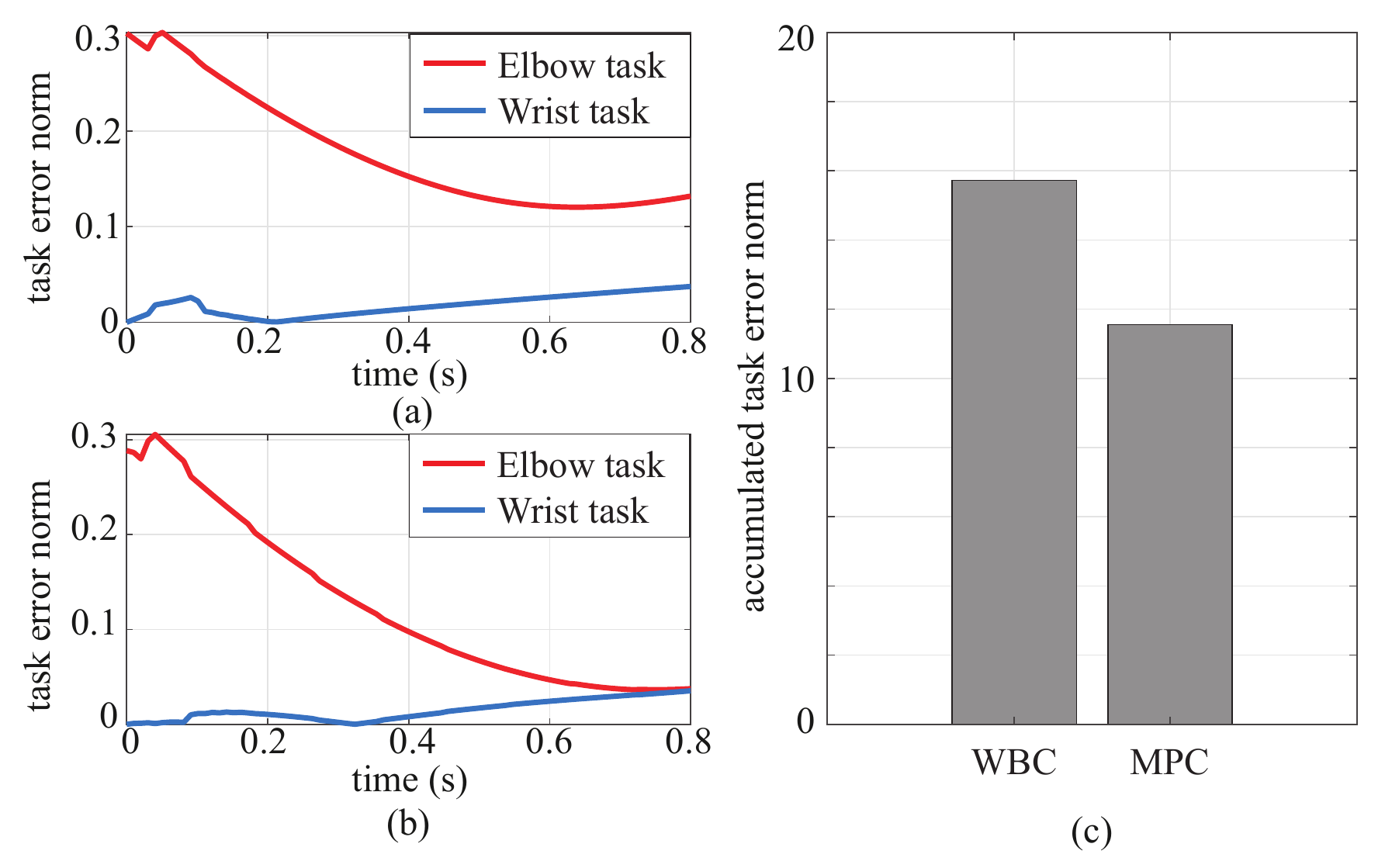}
\caption{Task error comparison over the time horizon $\mathbf{T}_N$: (a) task error norms when applying WBC, (b) task error norms when applying the proposed MPC, (c) accumulated task error norms over $\mathbf{T}_N$.}
\label{Fig5}
\end{figure}

The position errors of both tasks are shown in Fig. \ref{Fig3} and Fig. \ref{Fig4}. The maximum errors of the wrist position driven by WBC and the proposed MPC are $[0.0177, \: 0.0375]$ and $[0.0208,\: 0.0413]$, respectively. For the elbow positioning task, both control approach produce the maximum errors $[0.1310, \: 0.3033]$ and $[0.0577, \: 0.3057]$ respectively. We also compute the norm of each error to show that the defined task hierarchy is valid in these numerical simulations. Fig. \ref{Fig5} shows the error norms of all tasks over the finite-time horizon $\mathbf{T}_N$. The error norms for the wrist positioning task is smaller than those for the elbow positioning task over $\mathbf{T}_{N}$. Also, we accumulated the error norms, which are $15.7235$ and $11.5531$, and compare them with each other in Fig. \ref{Fig5}(c). The proposed MPC-based control approach obtains smaller task error and keeps the defined hierarchy over the finite-time horizon. 

\section{Conclusion}
This paper proposes a control approach for executing multiple hierarchical tasks on underactuated and constrained robots. To the best of our knowledge, this paper is the first one to implement WBC to constrained and underactuated robots executing hierarchical tasks within the framework of (convex) MPC. Conventional WBCs and OSCs generate instantaneously optimal (myopic) solutions which are not optimal over longer time horizons. However, the proposed control approach can obtain recursively optimal solutions over finite time horizons. Another contribution of this paper is the formulation of quadratic constraints that reflect the hierarchy of tasks assigned to the robots. Compared to WBC, the proposed MPC-based method reduces significantly the sum of errors for all tasks over the full time horizon.

Our extensive numerical simulations have shown that the computational time can be significantly reduced by linearizing the state equation and by convexifying all costs and constraint functions. In future work, we will analyze the computational cost of the algorithm in more detail and we will propose ways to reduce it. Furthermore, we will validate the proposed method through numerous (real) experiments using real robots like \textit{Scorpio}. Furthermore, we will extend our approach for the case of robotic systems operating in uncertain (stochastic) environments (subject to, for instance, stochastic disturbances) by employing stochastic MPC techniques.

\section*{ACKNOWLEDGMENT}
The authors would like to thank the members of the Human Centered Robotics Laboratory at The University of Texas at Austin and Apptronik Systems, Inc. for their great help and support. 

\bibliographystyle{IEEEtran}
\bibliography{ref}

\end{document}